\documentclass[11pt]{article}
\usepackage{coling2018}
\usepackage{times}
\usepackage{url}
\usepackage{latexsym}
\usepackage{mathrsfs}
\usepackage{amsmath}
\usepackage{amssymb}
\usepackage[margin=1in]{geometry}

\title{Comparing morphological complexity of Spanish, Otomi and Nahuatl}

\author{Ximena Gutierrez-Vasques \\
Universidad Nacional Aut\'onoma \\ de M\'exico \\
Mexico City\\
  {\tt xim@unam.mx} \\\And
  Victor Mijangos \\
Universidad Nacional  Aut\'onoma \\ de M\'exico\\
  Mexico City\\
  {\tt vmijangosc@ciencias.unam.mx} \\}

\date{}

\begin{document}
\maketitle
\begin{abstract}
We use two small parallel corpora for comparing the morphological complexity of Spanish, Otomi and Nahuatl. These are languages that belong to different linguistic families, the latter are low-resourced. We take into account two quantitative criteria, on one hand the distribution of types over tokens in a corpus, on the other, perplexity and entropy as indicators of word structure predictability. We show that a language can be complex in terms of how many different morphological word forms can produce, however, it may be less complex in terms of predictability of its internal structure of words.

\end{abstract}

%\section{Stuff by bazofies} YA LO AGREGUÉ ABAJO

%In general terms, two point of view can be distinguished \cite{miestamo2008grammatical}: the absolute, which defines complexity in terms of the number of parts of a system; and the relative, which refers to terms of cost and difficulty to language users. The later, however, has been considered as an approach apart from complexity \cite{dahl2004growth} in the sense that refers to subjective perspectives if the speakers. For this reason, some authors \cite{miestamo2008grammatical,dahl2004growth,baerman2015understanding} prefers, when speaking of complexity, to refer only to the absolute approach that is theory-oriented and, then, more objective.

%Another common distinction in language complexity studies is that between global and particular complexity. Global complexity seeks for characterize entire languages as easy or difficult to learn \cite[p. 29]{miestamo2008grammatical}, while particular complexity refers only to a level of the whole language (for example phonological complexity, morphological complexity, syntactical complexity).

\section{Introduction}

Morphology deals with the internal structure of words \cite{aronoff2011morphology,haspelmath2013understanding}. Languages of the world have different word production processes. Morphological richness vary from language to language, depending on their linguistic typology. In natural language processing (NLP), taking into account the morphological complexity inherent to each language could be important for improving or adapting the existing methods, since the amount of semantic and grammatical information encoded at the word level, may vary significantly from language to language.

Conceptualizing and quantifying linguistic complexity is not an easy task, many quantitative and qualitative dimensions must be taken into account \cite{miestamo2008grammatical}. On one hand we can try to answer what is complexity in a language and which mechanisms express it, on the other hand, we can try to find out if there is a language with more complex phenomena (phonological, morphological, syntactical) than other and how can we measure it. 
\newcite{miestamo2008grammatical} distinguishes between two types of complexity: the absolute, which defines complexity in terms of the number of parts of a system; and the relative, which is related to the cost and difficulty faced by language users. Some authors focuses in the absolute approach since it is less subjective. Another common complexity distinction is between global and particular. Global complexity characterizes entire languages, e.g., as easy or difficult to learn \cite[p. 29]{miestamo2008grammatical}, while particular complexity refers only to a level of the whole language (for example phonological complexity, morphological complexity, syntactical complexity).

We focus on morphological complexity. Many definitions of this term have been proposed \cite{baerman2015understanding,anderson2015dimensions,sampson2009language}. From the computational linguistics perspective there has been a special interest in corpus based approaches to quantify it, i.e.,  methods that estimate the morphological complexity of a language directly from the production of morphological instances over a corpus. 
This type of approach usually represents a relatively easy and reproducible way to quantify complexity without the strict need of linguistic annotated data.
The underlying intuition of corpus based methods is that morphological complexity depends on the morphological system of a language, like its inflectional and derivational processes. A very productive system will produce a lot of different word forms. This morphological richness can be captured with several statistical measures, e.g., information theory measures \cite{blevins2013information} or type token relationships. For example, \newcite[p.~9]{bybee2010language} affirms that ``the token frequency of certain items in constructions [i.e., words] as well as the range of types [...] determines representation of the construction as well as its productivity''.

In this work, we are interested in using corpus based approaches; however, we would like to quantify the complexity not only by the type and token distributions over a corpus, but also by taking into account other important dimension: the predictability of a morph sequence \cite{montermini2013stem}. %\footnote{We understand `morph' as recurrent and minimal common sub-string from a set of words in systematic co-variation \cite{montermini2013stem}. This term is different from `morpheme' in the sense that it is not necessary associated with a morphosyntactic feature.}. 
This is a preliminary work that takes as a case of study the distant languages Otomi, Nahuatl and Spanish. The general idea is to use parallel corpora, type-token relationship and some NLP strategies for measuring the predictability in statistical language models. 

Additionally, most of the previous works do not analyze how the complexity changes when different types of morphological normalization procedures are applied to a language, e.g., lemmatization, stemming, morphological segmentation. This information could be useful for linguistic analysis and for measuring the impact of different word form normalization tools depending of the language. In this work, we analyze how the type-token relationship changes using different types of morphological normalization techniques.

\subsection{The type-token relationship (TTR)}
The type-token relationship (TTR) is the relationship that exists between the number of distinct words (types) and the total word count (tokens) within a text. This measure has been used for several purposes, e.g., as an indicator of vocabulary richness and style of an author \cite{herdan1966advanced,stamatatos2009survey}, information flow of a text \cite{altmann2008anleitung} and it has also been used in child language acquisition, psychiatry and literary studies \cite{malvern2002investigating,kao2012computational}.

TTR has proven to be a simple, yet effective, way to quantify the morphological complexity of a language. This is why it has been used to estimate morphological complexity using relatively small corpora \cite{kettunen2014can}. It has also shown a high correlation with other types of complexity measures like entropy and paradigm-based approaches that are based on typological information databases \cite{bentz2016comparison}

It is important to notice that the value of TTR is affected by the type and length of the texts. However, one natural way to make TTRs comparable between languages is to use a parallel corpus, since the same meaning and functions are, more or less, expressed in the two languages. When TTR is  measured over a parallel corpus, it provides a useful way to compare typological and morphological characteristics of languages. \newcite{kelih2010type} works with parallel texts of the Slavic language family to analyze morphological and typological features of the languages, i.e., he uses TTR for comparing the morphological productivity and the degree of syntheticity and analycity between the languages. Along the same line, ~\newcite{mayer2014extraction} automatically extract typological features of the languages, e.g., morphological synthesis degree, by using TTR.

There exist several models that have been developed to examine the relationship between the types and tokens within a text \cite{mitchell2015type}. The most common one is the ratio $\frac{types}{tokens}$ and it is the one that we use in this work.

\subsection{Entropy and Perplexity}
In NLP, statistical language models are a useful tool for calculating the probability of any sequence of words in a language. These models need a corpus as training data, they are usually based on n-grams, and more recently, in neural representations of words.

Information theory based measures can be used to estimate the predictiveness  of these models, i.e., perplexity and entropy. Perplexity is a common measure  for the complexity of n-grams models in NLP \cite{brown1992class}. Perplexity is based in Shannon's entropy \cite{shannon1951mathematical} as the perplexity of a model $\mu$ is defined by the equation $2^{H(\mu)}$, where $H(\mu)$ es the entropy of the model (or random variable). Shannon's entropy had been used for measuring complexity of different systems. In linguistics, entropy is commonly used to measure the complexity of morphological systems \cite{blevins2013information,ackerman2013morphological,baerman2012paradigmatic}. Higher values of perplexity and entropy mean less predictability.

Perplexity depends on how the model is represented (this includes the size of the data). In this work, we compare two different models for calculating the entropy and perplexity: a typical bigram model adapted to a morph level\cite{brown1992class}; and our proposal based on using the word as a context instead of ngrams.
 
We rely in parallel corpora to compare the measures across languages, since the same meaning and functions are shared in the two languages.

\begin{description}

    \item[Bigram model.] This model takes into consideration bigrams \cite{brown1992class} as context for determining the joint probabilities of the sub-strings. Here the bigrams are sequences of two morphs in the text (whether they belong to the same word or not). This is a typical statistical language model but instead of using sequences of words, we use morphological segmented texts. In addition, we use a Laplacian (or add one) smoothing for the conditional probabilities \cite{chen1999empirical}.

%    \item[Neural model.] For the neural model, we train the sub-strings with Word2Vec to obtain vector representations of each sub-string (morph) in a word context. Then, we determine the conditional probability with the Softmax function:
    
 %   \begin{equation}
 %       p(x|y) = \frac{exp(-x^Ty)}{\sum_z exp(-z^Ty)}
 %   \end{equation}
    
 %   Were $x,y \in \mathbb{R}^n$ are vectors representing morphs.
    
    \item[Word level.] The word level representation takes the whole word as context for the determination of joint probabilities. Therefore, the frequency of co-occurrence is different from zero only if the sub-word units (morphs) are part of the same word. For example, if $xby$ is a word with a prefix $x$ and a suffix $y$, the co-occurrence of $x$ with $b$ will be different from zero as both morphs are part of the word $xby$. Similarly, the co-occurrence of $y$ with $b$ will be different from zero. Conversely, if two morphs are sub-strings of different words, its co-occurrence will be zero.
    To calculate the conditional probabilities we use and add one estimator defined as:
    
    \begin{equation}
        p(x|y) = \frac{fr(x,y) + 1 }{fr(x,y) + V}
    \end{equation}
     Where $V$ is the number of types and $fr(\cdot)$ is the frequency of co-occurrence function.

\end{description}

\section{Experimental setting}
\subsection{The corpus}
We work with two language pairs that are spoken in the same country (Mexico) but they are typologically distant languages: Spanish (Indo-European)-Nahuatl (Uto-Aztecan) and Spanish-Otomi (Oto-Manguean). Both, Nahuatl and Otomi are low-resource languages that face scarcity of digital parallel and monolingual corpora.

Nahuatl is an indigenous language with agglutinative and polysynthethic morphological phenomena. It can agglutinate many different prefixes and suffixes to build complex words. Spanish also has rich morphology, but it mainly uses suffixes and it can have a fusional behavior, where morphemes can be fused or overlaid into a single one that encodes several grammatical meanings. Regarding to Otomi, its morphology also has a fusional tendency, and it is head-marking. Otomi morphology is usually considered quite complex \cite{palancar2012conjugation} as it exhibits different phenomena like stem alternation, inflectional class changes and suprasegmental variation, just to mention some.

Since we are dealing with low resource languages that have a lot of dialectal and orthographic variation, it is difficult to obtain a standard big parallel corpus. We work with two different parallel corpora, i.e., Spanish-Nahuatl and Spanish-Otomi. Therefore the complexity comparisons are always in reference to Spanish.

We used a Spanish-Nahuatl parallel corpus created by \newcite{GUTIERREZVASQUES16.1068}. However, we used only a subset since the whole corpus is not homogeneous, i.e., it comprises several Nahuatl dialects, sources, periods of time and it lacks of a general orthographic normalization. We chose the texts that had a more or less systematic writing. On the other hand, we used a Spanish-Otomi parallel corpus \cite{lastra1992otomi} conformed by 38 texts transcribed from speech. This corpus was obtained in San Andr\'es Cuexcontitlan. It is principally composed by narrative texts, but also counts with dialogues and elicited data. 
Table \ref{parallelcorpus} shows the size of the parallel corpora used for the experiments.

\begin{table}[!h]
\small
\begin{center}
\begin{tabular}{|l|l|l|}
\hline \bf Parallel Corpus & \bf Tokens & \bf Types \\ \hline
\bf Spanish-Nahuatl & &  \\ \hline
Spanish (ES) & 118364 & 13233 \\ \hline
Nahuatl (NA) & 81850 & 21207 \\ 
\hline 
\bf Spanish-Otomi & &  \\ \hline
Spanish (ES) & 8267 & 2516 \\
\hline
Otomi (OT) & 6791 & 3381 \\
\hline

\end{tabular}
\end{center}
\caption{\label{parallelcorpus} Size of the parallel corpus }
\end{table}
%\begin{table}[h]
%\begin{center}
%\begin{tabular}{|c|c|c|c}
%\hline \bf ES & \bf Tokens & \bf Types \\ \hline
%Training & 2175533 & 99564 \\
%\hline Development & 800 & 800 \\
%\hline Test & 792 & 792 \\
%\hline \bf NA  & \multicolumn{2}{|c|}{}  \\
%\hline Training & 83229 & 22174 \\
%\hline Development & 1379 & 1379 \\
%\hline Test & 288 & 288 \\
%\hline
%\end{tabular}
%\end{center}
%\caption{\label{monolingualcorpus} Size of monolingual corpora used to generate morphological segmentation models}
%\end{table}
%In addition to the parallel corpus, we incorporated monolingual texts, in order to generate unsupervised morphological segmentation models that are explained in the section \ref{secc:morpho}.

%\textbf{VICLTES} %incluido
%Otomi is an indigenous language spoken in Mexico. It is part of the Oto-Mangue family.  Typlogically, its morphology is fusional, and head-marking. Otomi morphology is considered very complex \cite{palancar2012conjugation} as it presents different phenomena like stem alternation, inflectional class changes, suprasegmental variation and so on.
\subsection{Morphological analysis tools}
\label{secc:morpho}
We used different morphological analysis tools, in order to explore the morphological complexity variation among languages and between the different types of morphological representations. We performed lemmatization for Spanish language, and morphological segmentation for all languages.

In NLP, morphology is usually tackled by building morphological analysis (taggers) tools. And more commonly, lemmatization and stemming methods are used to reduce the morphological variation by converting words forms to a standard form, i.e., a lemma or a stem. However, most of these technologies are focused in a reduced set of languages. For languages like English, with plenty of resources and relatively poor morphology, morphological processing may be considered solved.

However, this is not the case for all the languages. Specially for languages with rich morphological phenomena where it is not enough to remove inflectional endings in order to obtain a stem.

Lemmatization and stemming aim to remove inflectional endings. Spanish has available tools to perform this task. We used the tool Freeling\footnote{http://nlp.lsi.upc.edu/freeling/}. Regarding to morphological segmentation, we used semi-supervised statistical segmentation models obtained with the tool Morfessor \cite{virpioja2013morfessor}. In particular, we used the same segmentation models reported in \newcite{ximena2017bilingual} for Spanish and Nahuatl. As for Otomi, we used manual morphological segmentation of the corpus, provided by a specialist.

\subsection{Complexity measures}

We calculated the type-token relationship for every language in each parallel corpus. Table \ref{TTR1} shows the TTR of the texts without any processing ($ES$, $NA$) and with the different types of morphological processing: morphological segmentation ($ES_{morph}$, $NA_{morph}$), lemmatization ($ES_{lemma}$). In a similar way, Table \ref{TTR2} shows the TTR values for the Spanish-Otomi corpus. It is worth mentioning that the TTR values are only comparable within the same parallel corpus.

\begin{table}[!h]
\small
\begin{center}
\begin{tabular}{c c c c}
\hline ~ & \bf Tokens & \bf Types & \bf TTR ($\%$) \\ \hline
$ES$ & 118364 & 13233 & 11.17 \\ 
$NA$ & 81850	& 21207	& \bf 25.90\\ \hline
$ES_{morph}$ & 189888	& 4369 & 2.30 \\ 
$NA_{morph}$ & 175744	& 2191 & \bf 1.24 \\ \hline
$ES_{lemma}$ & 118364 & 7599 & 6.42  \\ \hline
\end{tabular}
\end{center}
\caption{\label{TTR1} TTR for Nahuatl-Spanish corpus}
\end{table}

\begin{table}[!h]
\small
\begin{center}
\begin{tabular}{c c c c }
\hline ~ & \bf Tokens & \bf Types & \bf TTR ($\%$) \\ \hline
$ES$ & 8267 & 2516 & 30.43 \\ 
$OT$ & 6791 & 3381	& \bf 49.78 \\ \hline
$ES_{morph}$ & 14422 & 1072 & 7.43 \\ 
$OT_{morph}$ & 13895	& 1788 & \bf 1.28 \\ \hline
$ES_{lemma}$ & 8502 & 1020 & 8.33\\ \hline
\end{tabular}
\end{center}
\caption{\label{TTR2} TTR for Otomi-Spanish corpus}
\end{table}

We also calculate the perplexity and complexity for the different languages. Since we are focusing on morphological complexity, we took only the segmented data for computing the entropy and the perplexity. We do not use the lemmatized or non segmented data since this would be equivalent to measuring the combinatorial complexity between words, i.e. syntax. In this sense, the entropy and perplexity reflects the predictability of the morphs sequences. Tables ~\ref{tab:perplexity} and  \ref{tab:entropy} shows the perplexity and entropy in each language pair.

%caption{\labl{TTRrocessing2} TTR difference
%We also calculate TTRs using the segmentation models that had the poorest performance ($morph2$) for both languages. Tables \ref{TTR11}, \ref{TTR22}.
%\begin{table}[h]
%\begin{center}
%\begin{tabular}{|c|c|c|c|}
%\hline \bf  & \bf Tokens & \bf Types & \bf TTR ($\%$) \\ \hline
%$ES_{morph2}$ & 172288	& 9878 & 5.73 \\ \hline
%$NA_{morph2}$ & 117656	& 16136 &13.71 \\ \hline
%\end{tabular}
%\end{center}
%\caption{\label{TTR11} TTR for different types of morphological processing}
%\end{table}
%\begin{table}[h]
%\begin{center}
%\begin{tabular}{|l|c|}
%\hline \bf  & \bf Diff TTR ($\%$)\\ \hline
%$ES-NA_{morph2}$ & 2.53 \\ \hline
%$ES_{stem}-NA_{morph2}$ & 6.74 \\ \hline
%$ES_{lemma}-NA_{morph2}$ & 7.29 \\ \hline
%$ES_{morph2}-NA_{morph2}$ & 7.98 \\ \hline
%$ES_{morph2}-NA$ & 20.17 \\ \hline
%\end{tabular}
%\end{center}
%\caption{\label{TTR22}TTR difference between languages}
%\end{table}

%%%%
%SOLVED
%{Victli: A partir de aqui todo se vuelve un desastre, observaciones :
%1. hay qu enormalizar teminologia, que es eso desegm ve las tabalas de arriba veniamos manejando morph  subindices y todo %eso
%2. Por aue un corpus si tiene elespañol segmentado y el otro %no? si lo pusimosen uno lo debemos poner en el otro. Ademas %quenoya habiamos calculado todo eso en bilingual abejita?
%3. por que aqui no aparecen lostextos normales sin %normalizacion morfologica? porque no tiene sentido? si este es %el casohay que explicarlo, porque en TTr si pusimos es na %normal, para queno se confunda el lector}

 \begin{table}[!h]
\small
    \centering
    \begin{tabular}{c c c} \\ \hline
            ~ & \textbf{Word level} & \textbf{Bigram model}  \\ \hline %& \textbf{Neural model} \\ \hline
            ~ &  ES-NA & ~ \\ \hline
        $NA_{morph}$ & 214.166 & 1069.973 \\ %& 1701.321 \\
         $ES_{morph}$ & 1222.956 & 2089.774 \\ %& 3189.915 \\
         %Spanish (Lem) & ~ & 5777.862 & 7595.599 \\ 
         \hline
         ~ & ES-OT & ~  \\ \hline
         $ES_{morph}$ & 208.582 & 855.1766 \\ %& 1070.999  \\
          $OT_{morph}$ & 473.830 & 1315.006 \\ %& 1481.999 \\ 
          \hline
    \end{tabular}
    \caption{\small Perplexity obtained in the different parallel corpora}
    \label{tab:perplexity}
\end{table}

\begin{table}[!h]
\small
    \centering
    \begin{tabular}{c c c} \\ \hline
           ~ & \textbf{Word level} & \textbf{Bigram model}  \\ \hline%& \textbf{Neural model} \\ \hline
            ~ & ES-NA & ~ \\ \hline
         $NA_{morph}$ & 0.697 & 0.906 \\%& 0.966\\
         $ES_{morph}$ & 0.848 & 0.911 \\ \hline %& 0.962 \\ \hline
         %Spanish (Lem) & ~ & 0.969 & 0.999 \\ \hline
         ~ &  ES-OT & ~  \\ \hline
        $ES_{morph}$  & 0.765 & 0.967 \\%& 0.999 \\
         $OT_{morph}$  & 0.843 & 0.984 \\ \hline %& 0.999 \\
    \end{tabular}
    \caption{ \small Entropy obtained in the different parallel corpora}
    \label{tab:entropy}
\end{table}

\section{Results analysis}
\subsection{TTR as a measure of morphological complexity}
%morph 2 maintain the reduction, however when compared the differences does not make many sense
When no morphological processing is applied, Nahuatl has a lot higher TTR value than Spanish, i.e., a greater proportion of different word forms (types). In spite of Nahuatl having fewer tokens because of its agglutinative nature, it has a lot more types than Spanish. This suggests that Nahuatl has a highly productive system that can generate a great number of different morphological forms. In other words, it is more likely to find a repeated word in Spanish than in a Nahuatl corpus.
In the case of Otomi-Spanish, Otomi also has a bigger complexity compared to Spanish in terms of TTR. Even though both Otomi and Spanish show fusional patterns in its inflection, Otomi also count with a lot of derivational processes and shows regular stem alternations. 

%when some type of morphological process was applied to the texts, TTR decreased, which can be interpreted in terms of reduction of the morphological complexity of the doff

In every case, morphological segmentation induced the smallest values of TTR for all languages. Suggesting that greater reduction of the morphological complexity is achieved when the words are split into morphs, making it more likely to find a repeated item. For instance, when Nahuatl was morphologically segmented, TTR had a dramatic decrease (from $26.22$ to $1.23$). This TTR reduction could be the result of eliminating the combinatorial variety of the agglutinative and polysynthetical morphology of the language. Therefore, when we segment the text we break this agglutination, leading to significantly less diverse units.

In the case of Otomi language, a similar trend can be observed. Otomi seems to be morphologically more complex than Spanish in terms of TTR, i.e., more diverse types or word forms. When morphological segmentation is applied, TTR decreases and Otomi language has a lower TTR compared to Spanish. Even though Otomi is not a polysynthetic language like Nahuatl, these results suggest that Otomi has also a great combinatory potential of its morphs, i.e, when Otomi gets morphologically segmented we obtain less diverse types, these morphs may be recurrent in the text but they can be combined in many several ways within the Otomi word structure. Linguistic studies have shown that Otomi language can concatenate several affixes, specially in derivative processes \cite{lastra1992otomi}.

It has brought to our attention that  Spanish has a higher TTR than Nahuatl and Otomi, only when the languages are morphologically segmented. It seems that the morphs inventory is bigger in Spanish, we conjecture this is related to the fact that Spanish has more suppletion or ``irregular'' forms phenomena \cite{boye2006structure}.

\subsection{Predictability}
The predictability of the internal structure of word is other dimension of complexity. It reflects the difficulty of producing novel words given a set of lexical items (stems, suffixes or morphs).
First of all, as a general overview, we can see that word level models have the lower perplexity and entropy (Tables \ref{tab:perplexity} and \ref{tab:entropy}). We believe that this type of models capture better the morphological structure, since they take into account the possible combinations of morphs within a word and not outside the bounds of it (like the bigram model).

%Table~\ref{tab:entropy} and Table~\ref{tab:perplexity} shows that the word level models had the lower perplexity and entropy. This is not strange because this representation is the one that captures the morphological structure better as it takes the possible combinations between the morphs into a word and not outside the bounds out of it. The bigram model shows less clear results as it takes into consideration bounds that are outside the limits of the word.

It is interesting to compare the TTR and the predictability measures for each language. In the case of Nahuatl, TTR shows that there is a lot of complexity at lexical level (many different word forms, few repetitions), however, this contrasts with the predictability of the elements that conform a lexical item: the combination of morphs within a word is more predictable than Spanish, since it obtains lower values of Perplexity and entropy. The combinatorial structure of Nahuatl morphology shows less uncertainty than Spanish one, despite the fact that Nahuatl is capable of producing many more different types in the corpus due to its agglutinative and polysynthetic nature.

The case of Otomi language is different, since it seems that it is not only complex in terms of TTR but also in terms of predictability. It obtains higher entropy and perplexity than Spanish. We conjecture this is related to several phenomena. For instance, Otomi and Nahuatl allow a large number of morphs combinations to modify a stem (inflectional and derivational). However, Otomi shows phenomena that is not easy to predict; for example, it has a complex system of inflectional classes, stem alternations and prefix changes. Moreover, tones and prosody plays an important role in the morphology of Otomi verbs \cite{palancar2004verbal,palancar2016typology}. Also, we mentioned before that many of the affixes concatenations in Otomi take place in derivative processes. 
%Derivation tends to be less predictable than inflection phenomena, as derivation is less frequent and regular than inflection, and this could be an additional reason of why the entropy values of this language are high.
Derivation tends to be less predictable than inflection phenomena (derivation is less frequent and less regular), and this could be an additional reason of why the entropy values of this language are high.

%shows the highest entropy level. So, Otomi language show high complexity in both lexical level (TTR) and in the prediction of the internal structure of words. As Nahuatl, Otomi has a large number of morphs than can modify a stem (inflectional and derivational). However, Otomi shows phenomena that is not easy to predict; for example, it counts with a complex system of inflectional classes, stem alternations and preffix changes. Even more, it has been proved that tones and prosody plays an important role in the morphology of Otomi verbs \cite{palancar2004verbal,palancar2016typology}.

%comentar que está pasando pr ahi con el otomin y su contraste con TTR

\section{Conclusions}
%La conclusión puede ser solo un pequeño recuento con elementos de la introduccion y los haallazgos obtenidosidos

In this work we used corpus based measures like TTR, entropy and perplexity for exploring the morphological complexity of three languages, using two small parallel corpora. We use TTR as a measure of morphological  productivity of a language, and we use the entropy and perplexity calculated over a sequence of morphs, as a measure of predictability.

There may be a common believe that polysynthetical languages are far more complex than analytic ones. However, it is important to take into account the many factors that  lay a role in the complexity of the system. We stressed out that morphological complexity has several dimensions that must be taken into account \cite{baerman2015understanding}. 

While some agglutinative polysynthetical languages, like Nahuatl, could be considered complex by the number of morphemes the combinations and the information than can be encoded in a single word; the sequence of these elements may be more predictable than fusional languages like Spanish.

Languages like Otomi, showed high complexity in the two dimensions that we focused in this work (this is consistent with qualitative perspectives \cite{palancar2016typology}). 

These two dimensions of complexity are valid and complementary. Measures like TTR reflect the amount of information that words can encode in a language, languages that have a high TTR have the potential of encoding a lot of functions at the word level, therefore, they produce many different word forms. Perplexity and entropy measured over a sequence of morphs reflect the predictability or degree of uncertainty of these combinations. The higher the entropy (hence, the perplexity), the higher the uncertainty in the combinations of morphs.

%When talking of languages complexity, there exists a lot of preconceptions. For example, the common believe that polysynthetic languages are more complex than analytic ones. However, many factor plays a role for considering a language complex. Precise qualitative and quantitative methods need to be development in order to study complexity in linguistic terms.

This was a preliminary work. Deeper linguistic analysis, more corpora and more languages are needed. However, we believe that quantitative measures extracted from parallel corpora can complement and deepen the study of linguistic complexity.
Efforts are currently being made \cite{bane2008quantifying}. However, more studies are needed, especially for low resources languages.

\subsection{Future work}
Languages of the world have a wide range of functions that can be codified at the world level. Therefore, it would be interesting to consider the study of more complexity dimensions in our work. Popular quantitative approaches are successful in reflecting how many morphs can be combined into a single word. However, it is also important to take into account how complex the format of a word can be, i.e., not only how many elements can be combined but also what type of elements. For example, \newcite{dahl2009testing} argues that when a phoneme is added to a word, this process is not as complex as adding a tone.
%in a phonological word, adding a phoneme to manifest a morphological process is not such complex as adding a tone. 
%%%%%%VICTLE'S STUFF
%In other words, the status of the elements that combine to form words is a parameter for morphological complexity. 

Another interesting dimension is the complexity of the morphology in terms of acquisition (of native and L2 speakers). \newcite{miestamo2008grammatical} points out that this typo of complexity should be made on the basis of psycho-linguistics analysis in both processing and acquisition.

%Other dimension to be considered is how hard is the morphology of a language in terms of acquisition (native language and L2 language).  As is pointed by \newcite{miestamo2008grammatical} estimations about the complexity of a language for the different groups of languages users can only be made on the basis of psycho-linguistics analysis in both processing and acquisition.

Finally, one important factor that influences language complexity is culture. In many languages, pragmatics nuances are produced via morphological processes. For instance, languages like Nahuatl have a complex honorific or reverential system that is expressed using different types of affixes. Spanish expresses this type of phenomena with morphosyntactic processes. It is a challenging task to be able to quantify all these factors that play a role in the complexity of a language.

\section*{Acknowledgements}
This work was supported by the Mexican Council of Science and Technology (CONACYT), fund 2016-01-2225, and CB-2016/408885.
We also thank the reviewers for their valuable comments and to our friend Morris\'e P. Martinez for his unconditional support.

%The acknowledgements should go immediately before the references.  Do
%not number the acknowledgements section. Do not include this section
%when submitting your paper for review.

\bibliographystyle{acl}
\bibliography{acl2017}

\end{document}